\documentclass[runningheads]{llncs}

\usepackage{eccv}

\usepackage{eccvabbrv}

\usepackage{graphicx}
\usepackage{booktabs}
\usepackage[expansion=false]{microtype}

\usepackage[accsupp]{axessibility}  

\usepackage[pagebackref,breaklinks,colorlinks,citecolor=eccvblue]{hyperref}

\usepackage{orcidlink}

\usepackage{booktabs}
\usepackage{adjustbox}
\usepackage{subcaption}
\usepackage{soul,color}
\usepackage[dvipsnames]{xcolor}
\usepackage{microtype}
\usepackage{varwidth}

\begin{document}


\title{Human-Inspired Social Engagement Analysis via Interpretable Mutual Visual Attention} 

\titlerunning{Human-Inspired Social Engagement Analysis}

\author{Urwa Fatima\inst{1}\orcidlink{0009-0005-3948-1192} \and
Mohammad Zohaib \inst{1,2, 3}\orcidlink{0000-0003-2259-4121} \and
Francesca Odone \inst{1,2}\orcidlink{0000-0002-3463-2263} \and Nicoletta Noceti \inst{1,2} \orcidlink{0000-0002-6482-4768}}

\authorrunning{U.~Fatima et al.}

\institute{DIBRIS-Universit\`a degli Studi di Genova, Genova, IT \and
MaLGa - Machine Learning Genoa center, Genova, IT
\and
Istituto Italiano di Tecnologia, Genova, IT
}

\maketitle

\begin{abstract}
Understanding social interactions from non-verbal visual data is important for behavior analysis and activity monitoring.  We propose an interpretable computational model of social engagement inspired by psychological theories of mutual visual attention. Rather than learning interaction patterns end-to-end, our framework explicitly models dyadic visual attention and aggregates these cues into interpretable measures of individual and group engagement.
The resulting modular framework combines state-of-the-art  head orientation estimation with lightweight geometric reasoning, producing explanations that remain accessible to non-technical users.
We evaluate the proposed approach on a variety of data through quantitative experiments and demonstrate its practical usefulness with qualitative visualizations designed to support teachers, caregivers, and social workers in understanding group interaction dynamics.

\end{abstract}

\section{Introduction}
\label{sec:intro}


{
In the last decades, the importance of non-verbal signals in characterizing social interactions has become widely recognized. These subtle cues frequently convey critical information about human dynamics, emotions, and interpersonal relationships~\cite{ref:poria2019emotion,ref:rehg2013decoding,ref:stokoe2025categories,ref:mehrabian1971silent}. 
Psychological theories, particularly the Theory of Mind~\cite{ref:frith2012mechanisms}, highlight the fundamental role of non-verbal behaviours—especially mutual gaze and face-to-face attention—in conveying engagement and regulating social interactions. Despite this growing interest, computational approaches relying exclusively on non-verbal cues remain relatively limited, and are often restricted to solving individual tasks such as head pose estimation, body pose estimation, or action recognition~\cite{ref:diwan2025advancements,ref:guo2022mgtr,ref:guo2023social}.
}

{
In this work, we propose an interpretable and modular framework for video-based social interaction analysis that is grounded in a hierarchical model of human social reasoning \cite{kenny1984social,kenny2020dyadic}. According to this perspective, group engagement is not directly observed, but emerges from the network of interpersonal interactions among group members. Accordingly, our framework first models dyadic interactions from non-verbal visual cues and then aggregates these local relationships into interpretable measures of individual and group engagement \cite{argyle1994gaze,senju2009eye}.

More specifically, given a video of a small- or medium-sized group, the proposed pipeline progressively builds a multi-level representation of the social scene. Starting from head direction estimation, it infers pairwise visual attention and classifies dyadic interaction states, which are subsequently aggregated to compute individual- and group-level engagement by means of indices we designed to the purpose. The modular design makes each reasoning stage explicit and interpretable, enabling richer analysis and visualization than end-to-end approaches, where intermediate social cues are typically not explicitly represented.\\
We evaluate the proposed framework from both quantitative and qualitative perspectives. Quantitatively, we assess the representativeness of the pipeline in recognizing 
interaction patterns. Qualitatively, we show how the proposed engagement indices capture meaningful variations in group engagement across different interaction scenarios. To facilitate the interpretation of these results, we also introduce a set of visualization strategies that make the analysis accessible to users with little or no technical expertise.}
{
Indeed, the final goal of our research is to provide psychologists and educators, with easy to use tools to study social interactions in small groups. \\
%
For this we also carried out a live demonstration in a kindergarten, where our tool was used to visually describe children's engagement during a cooperative activity. Although the collected data can not be shared due to ethical constraints, teachers provided a very positive qualitative feedback on the interpretability and potential usefulness of the proposed visualizations.
}
%
%
%
To summarise, the main contributions of our work are the following:
{
\begin{itemize}
\item A modular video-based framework for hierarchical social interaction analysis, from pairwise attention estimation to group engagement.
\item Two engagement indices that quantify interaction at both the individual and group levels.
\item An intuitive visualization interface that makes multi-level social interaction analysis accessible to non-technical users.
\end{itemize}}


%
%
%

\section{Related Works}
\label{sec:literature_review}
%

{
A common principle in cognitively inspired approaches to social interaction analysis is that group-level behaviors emerge from pairwise interpersonal dynamics. This view is well established in social psychology through the Social Relations Model (SRM), which represents group interactions as the composition of actor, partner, and relationship effects, treating the dyad as the fundamental unit of social behavior \cite{kenny1984social, kenny2020dyadic}. More recent methodological frameworks extend this perspective to the analysis of groups, discussing hierarchical models in which group-level phenomena are inferred from dyadic interactions \cite{Kenny_Ackerman_Kashy_2024}. \\
This hierarchical assumption has also influenced human-centered computer vision.
In line with psychology literature, existing approaches view group engagement analysis as an emergent property arising from the network of social interactions among participants. In particular, Social Signal Processing assessed the use of gaze, head and body orientation, interpersonal distance, and conversational dynamics as key nonverbal cues for inferring social relationships and interaction patterns \cite{vinciarelli2009social, vinciarelli2011bridging}. Similarly, research on F-formations has demonstrated that these cues can be effectively combined to identify conversational partners and free-standing interaction groups \cite{cristani2013human, setti2015f}. More recently, comprehensive surveys on co-located human interaction analysis have highlighted hierarchical computational frameworks in which low-level behavioral cues are first integrated to estimate pairwise interpersonal relations, later exploited to infer higher-level group properties such as engagement, cohesion, rapport, and leadership \cite{beyan2023co,GROSSI202061}.\\
Of particular relevance for our work are approaches leveraging the concept of attention. In the past, mutual facing and shared attention have been exploited to detect dyadic engagement~\cite{ba2008recognizing,fathi2012social}, or to model group structure~\cite{setti2015f}. More recent deep learning approaches jointly reason over individuals and groups but require large datasets and lack interpretability~\cite{liu2019social,ref:kim2024modeling}. In our work, we leverage head direction as a proxy for gaze direction and attention ~\cite{ref:frith2012mechanisms,ref:algabri2024deep,ref:shepherd2010following,ref:grossmann2007development}.\\
Tightly coupled with social interaction is the concept of group engagement. Most previous works on engagement estimation focus on individuals. Head poses have been correlated with engagement in learning environments~\cite{zheng2025lightnet}, while dual-stream architectures combine facial and scene context~\cite{gothwal2025vibed}. In~\cite{ref:javed2020toward}, visual behavior and motion are used to estimate a continuous engagement for children with autism spectrum disorder. Transformer-based approaches have also been proposed to classify behavioral and collaborative engagement states using combined visual cues \cite{penchala2025learning}. Group engagement remains less explored, though recent work analyzes non-verbal behaviors as indicators of group engagement quality~\cite{noceti2025predicting,paneth2024zooming}.
}

\section{Proposed Framework}
\label{sec:proposed_framework}
\begin{figure*}[t]
    \centering
    \includegraphics[width=\linewidth]{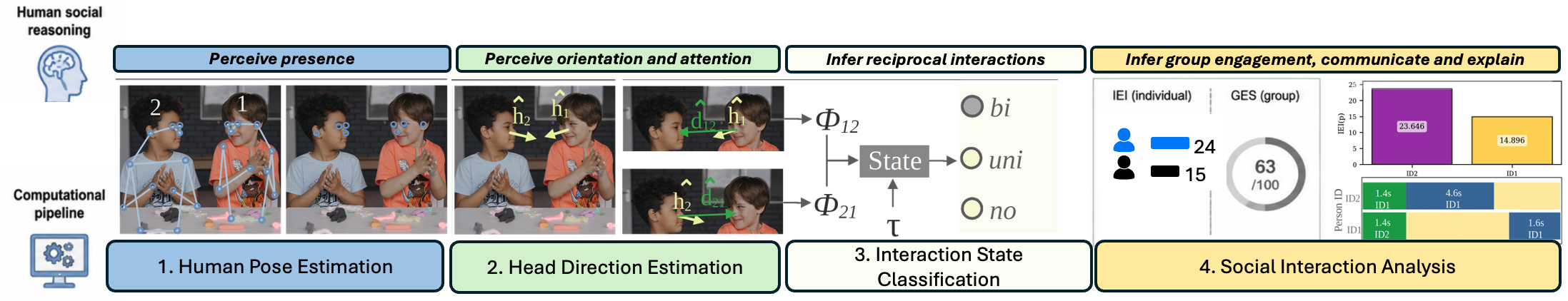}
    \caption{Proposed four-stage framework. 
    } 
    \label{fig:framework}
\end{figure*}

\noindent
{\bf Human-inspired design principle:} rather than directly predicting engagement from visual data, we mirror a hierarchical reasoning assessed in social cognition and interaction literature \cite{ref:frith2012mechanisms,kenny1984social, kenny2020dyadic}. Our design follows three principles: (i) attention is an observable cue from which engagement is inferred; (ii) reciprocal attention provides stronger evidence of interaction than one-sided attention; and (iii) group engagement emerges from the composition of local dyadic interactions. These principles lead to the design of a four stages framework, shown in  Fig.~\ref{fig:framework}.

\noindent
\textbf{Stage 1: Human Pose Estimation: }
Given an image, we extract each person’s pose and assign a unique ID for tracking, using YOLOv11 with ByteTrack for efficient multi-person pose estimation and tracking under occlusions. We retain five facial keypoints: the nose, left eye, right eye, left ear, and right ear, each one with its own confidence, setting missing points to (0, 0) with zero confidence.\\
%
%
%
%
\noindent
\textbf{Stage 2: Head Direction Estimation:}
Given the set of five facial 2D keypoints with confidence scores 
$\{(x_i, y_i, c_i)\}_{i=1}^{5}$ estimated in stage 1, stage 2 aims to estimate  head orientation as a triplet (yaw, pitch, roll). 
To the purpose, we first normalize facial keypoints to make them invariant to scale and absolute face position. Let $(\bar{x}, \bar{y})$ be the centroid of the facial keypoints; the normalized keypoints $\{(\tilde{x}_i, \tilde{y}_i, c_i)\}_{i=1}^{5}$ are computed as 
$\tilde{x}_i  = \frac{x_i - \bar{x}}{d_x^{\max}}$, and  $\tilde{y}_i = \frac{y_i - \bar{y}}{d_y^{\max}}$, 
where  $d_x^{\max}  = \max_i \left( |x_i - \bar{x}| \right)$ and $d_y^{\max} = \max_i \left( |y_i - \bar{y}| \right)$ are computed over keypoints with $c_i > 0$. 
Keypoints with zero confidence remain at $(0, 0)$ after normalization. 
The normalized keypoints are finally fed to HHP-Net ~\cite{ref:cantarini2022hhp,ref:tomenotti2024head} for the angles triplet prediction. 
%
As output, HHP-Net predicts the head orientation angles $(\theta, \alpha, \gamma)$ along with their respective uncertainties $(\sigma_\theta, \sigma_\alpha, \sigma_\gamma)$. 
These angles are used to generate a 2D unit vector $\vec{\hat{h}}$ on the image plane, according to the Tait-Bryan angles. The head direction is first represented as a vector $\vec h=(h_x, h_y)$ in pixel/image coordinates,  as  $h_x = \sin(\theta)$  and    $ h_y = -\cos(\theta)  \sin(\alpha) $
and then normalized to the unit norm as $\vec{\hat{h}} = \frac{\vec{h}}{\lVert \vec{h} \rVert}$.
The keypoint centroid $(\bar{x}, \bar{y})$ is used as the origin $\mathcal{O}$ of the head direction vector. This process is repeated for all $n$ people present in the image to obtain head direction vectors $\mathcal{H}= \{\vec{\hat{h}}_j\}_{j=1}^{n}$ and their origins $\mathcal{O} = \{\mathcal{O}_j\}_{j=1}^{n}$. 
These vectors are used in the next section to compute spatial relations between people.\\
\noindent \textbf{Stage 3: Interaction State Classification:}
Following prior work on mutual gaze~\cite{ref:gregorj2022influence,ref:lombardi2022toward}, interactions are categorized as bidirectional, unidirectional, or non-interacting state, corresponding to mutual gaze, one-sided gaze, and no gaze between individuals.
We detect these three interaction states among $n$ humans using head direction vectors $\vec{\hat{h}}$. We adopt a strategy presented in~\cite{ref:tomenotti2024head} which estimates the Looking At Each Other (LAEO) event (bidirectional state). We extend it to also identify unidirectional and non-interaction states as follows.\\
Consider a pair of individuals, $p_1$ and $p_2$, with head direction vectors and origins $(\vec{\hat{h}}_1, \mathcal{O}_1)$ and $(\vec{\hat{h}}_2, \mathcal{O}_2)$, respectively.
We compute their mutual visual attention in two steps. First, we compute the unit vectors between their head origins: $\vec{\hat{d}}_{12}$ points from $p_1$ to $p_2$  and is computed as $\frac{\mathcal{O}_2 - \mathcal{O}_1 }{\lVert \mathcal{O}_2 - \mathcal{O}_1 \rVert}$, while $\vec{\hat{d}}_{21}$ points from $p_2$ to $p_1$ and is simply the negative of $\vec{\hat{d}}_{12}$.
Second, we compute the angles $\phi_{12} = \arccos(\vec{\hat{d}}_{12} \cdot \vec{\hat{h}}_1),$ and $\phi_{21} = \arccos(\vec{\hat{d}}_{21} \cdot \vec{\hat{h}}_2)$ between each person’s head direction vector and the corresponding unit vector to the other person. 
These angles are used to measure mutual head-direction alignment and to classify interactions between $p_1$ and $p_2$. If both $\phi_{12}$ and $\phi_{21}$ are lower than a threshold $\tau$, the interaction is bidirectional; if only one of them is lower than $\tau$, it is unidirectional; otherwise it is non-interacting. The procedure is applied to all $\frac{n(n-1)}{2}$ pairs in the image.
This pairwise strategy does not guarantee consistency in the prediction. For instance, in a scene with only two individuals $p_1$ and $p_2$, having $p_1$ predicted as bidirectional and $p_2$ as unidirectional would be inconsistent.
Therefore, we first prioritize pairs classified as bidirectional interactions, assigning all members of these pairs to the bidirectional state. Next, we consider unidirectional interaction pairs and assign all remaining unassigned members to the unidirectional state, where the person whose angle is below $\tau$ is considered the looker and the other the target. Finally, any remaining members are classified as non-interacting.
This yields the interaction state of each person per frame, which is then applied sequentially across all video frames.\\
\noindent
\textbf{Stage 4: Social Interaction Analysis:}
We propose two engagement indices.
%
%

\noindent
\textit{Individual Engagement Index -- IEI($p$):} 
IEI represents each person’s participation in social interactions. For a given person $p$, it is computed from the number of frames spent in each interaction state with weights. 
Let $F_B$, $F_L$, and $F_T$ denote the frames where a person is classified as bidirectional, unidirectional looker, and unidirectional target. Then the IEI($p$) is computed as: 
%

\begin{equation}
\operatorname{IEI}(p)
=
\frac{100}{N}
\sum_{c\in\{B,L,T\}}
\omega_c F_c\
\label{eq:IEI}
\end{equation}
where $N$ is the  number of frames in which person p is detected, and $\omega_B$, $\omega_L$, and $\omega_T$ are weights for each interaction state in engagement analysis. We empirically set the weights to $\omega_B=1.00$, $\omega_L=0.30$, and $\omega_T=0.05$, respectively. Mutual events indicate the highest engagement and thus receive the highest weight. In unidirectional interactions, the looker is more engaged than the target, so $\omega_L > \omega_T$.
IEI($p$) ranges from 0 to 100, where 0 indicates a non-interacting person and 100 corresponds to a person remaining in the bidirectional state throughout the video. Comparing IEI($p$) across participants helps identify highly engaged individuals and differences in attention during the interaction.

%
%
%
%

\noindent
\textit{Group Engagement Score (GES):} It provides a group-level analysis of social interaction across the entire video.
Given the total number of people $n$ in each frame $t \in [1, N]$,  which we assume not to change over time,  the GES is computed as a weighted ratio between the predicted engagement $\hat{E}$ and the maximum possible engagement $E^{max}$ over a video of $N$ frames.  
For the $n$ subjects in a frame $t$, the maximum possible number of bidirectional $B^{max}_t$ and unidirectional $U^{max}_t$ states are $\left\lfloor n/2\right\rfloor$ and $n \% 2$, respectively.  
Let $\hat{B}_t$ and $\hat{U}_t$ denote the predicted number of bidirectional and unidirectional states in frame $t$.  $\beta_{\text{B}}=1$ and $\beta_{\text{U}}=0.5$ are their respective weights.  
Then

\begin{equation}
E^{\max}
=
\sum_{t=1}^{N}
\left(
\beta_{\mathrm{B}} B_t^{\max}
+
\beta_{\mathrm{U}} U_t^{\max}
\right),\hspace{0.5cm}
\hat{E}
=
\sum_{t=1}^{N}
\left(
\beta_{\mathrm{B}} \hat{B}_t
+
\beta_{\mathrm{U}} \hat{U}_t
\right).
\label{eq:GES}
\end{equation}

\noindent and 
$GES=100 \frac{\hat{E}}{E^{max}}$
if  $E^{max} > 0$, 
        $0$ otherwise. 
        A score of $0$ indicates no interaction, while $100$ indicates the maximum possible interaction.

\section{Experiments and Results}


\noindent {\bf Benchmarks. }{The experimental assessment aims to evaluate our approach in different environments and conditions and for this reason we consider several benchmarks. 
First we consider the {\em GP-Static++ }dataset ~\cite{ref:peng2025multi},which includes 71 videos of  dyadic interactions ($\sim$ 1.8 -- 17 sec). To the best of our knowledge, it is the only available dataset with appropriate annotations. On this dataset we provide a quantitative assessment of interaction classification. 
In addition, we consider 3 triadic conversational scenes selected from the {\em CMU Panoptic Haggling} dataset~\cite{ref:Joo_2019_CVPR}, where three participants engage in face-to-face verbal interaction ($\sim$ 4 -- 6 min). The sequences have been manually annotated according to the annotation of GP-Static++, to allow a comparison of performance for dyads and triplets. 
For a qualitative evaluation of visualizations and engagement indices, we also rely on an additional data source targeting high-resolution group activities across diverse ages and indoor/outdoor settings. To this end we adopted the publicly available unannotated {\em Pexels} videos\footnote{https://www.pexels.com/, Accessed: June. 2026} (15 videos, $\sim$ 0.15 -- 2 min).} \\
\noindent {\bf Implementation Details. }We used the yolo11n-pose variant of YOLOv11 with ByteTrack. We set the detection and IoU threshold of YOLO to 0.5. The  
match and track thresholds of the tracker are set to 0.8 and 0.5, respectively. We empirically select for $\tau$ the value $\pm 15^\circ$.
We used an official TensorFlow implementation of HHP-Net with pretrained checkpoints, without any additional fine-tuning.
All experiments were performed on an ASUS TUF Gaming F15 laptop with an Intel Core i7 processor and 16 GB RAM. \\
\noindent
{\bf Quantitative Results  (Tab.~\ref{tab:overall_f1_by_type}). } We compare our framework with the frame-based method in \cite{ref:peng2025multi} on the GP-Static++ dataset. We provide the global F1 score on the test set of GP-Static++, together with the performance for each interaction class. Our method provides superior performances in all scenarios. It is also worth mentioning that, differently from \cite{ref:peng2025multi}, our framework is training-free and does not rely on any task-specific training data. \\
In addition, we evaluate the prediction of our method on the selection of annotated sequences from the CMU dataset. This allows us to provide an evidence of the influence on the number of people in the group.  From the F1 scores, we may observe that our method appears to be influenced by the cardinality of the group regarding the correct classification of unidirectional or no interaction, while only a small degradation is observed for mutual events, likely because of the robustness thanks to the stronger geometric constraints.

\begin{table}[!t]
    \centering
    \caption{
    Comparison with the frame-based method  \cite{ref:peng2025multi} on the GP-static++ dataset. In addition to global F1 score on the test set, we also provide the performance of bidirectional events ($F1_B$), unidirectional events ($F1_U$), and no interaction ($F1_{NO}$).
    We also report results on the CMU dataset showing the influence of group cardinality (2 for GP-Static++, 3 for CMU).
    }
    \label{tab:overall_f1_by_type}
    \begin{tabular}{lclccccc}
        \toprule
        Dataset & Frames $ \ \ $ & Method $ \ \ $ & F1$_{B} \uparrow$ & F1$_{U} \uparrow$ & F1$_{NO} \uparrow$ & Avg. F1 $\uparrow$ \\
        \midrule
        GP-Static++  &  13789  $ \ \ $ & \cite{ref:peng2025multi} &  0.56  & 0.44  &  0.49  & 0.49 \\
          &  &  	Ours &  \textbf{0.70}  & \textbf{0.45}  & \textbf{0.60} &  \textbf{0.58} \\
        \hline  
        CMU  & 2368  $ \ \ $   & 	Ours & 0.68 & 	0.29 & 	0.54 & 	0.50 \\
     
        \bottomrule
    \end{tabular}
\end{table}
%
%
%




\begin{figure*}[!t]
    \centering
    \begin{subfigure}{\textwidth}
        \centering
        \includegraphics[width=.82\linewidth]{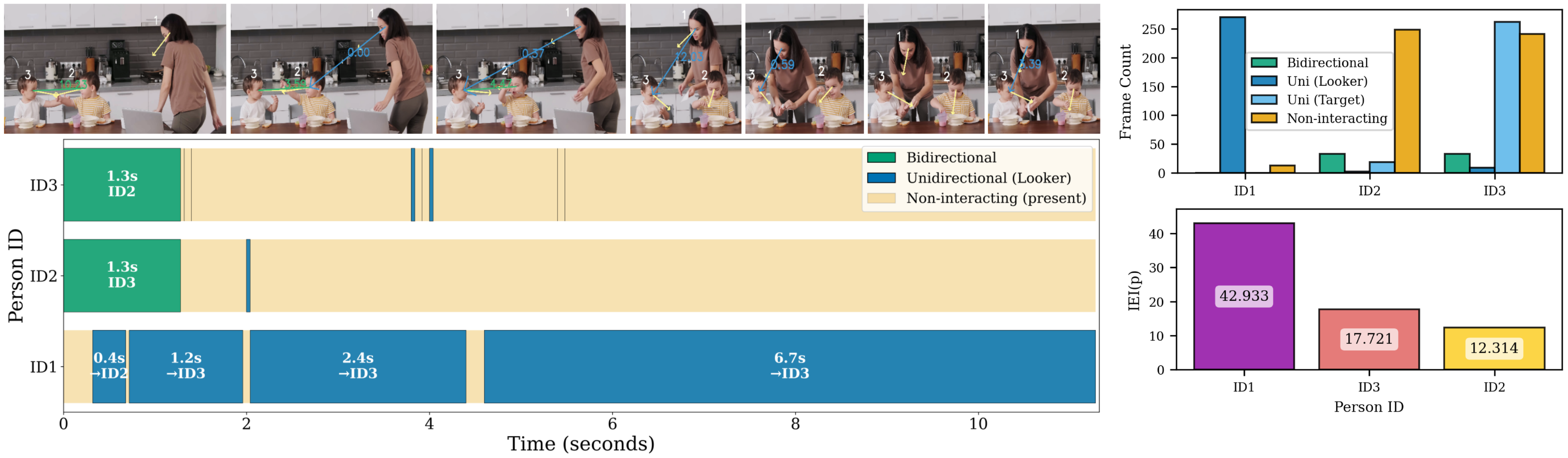}
        \caption{}
        \label{fig:sub1}
    \end{subfigure}
    \begin{subfigure}{\textwidth}
         \centering
         \includegraphics[width=.82\linewidth]{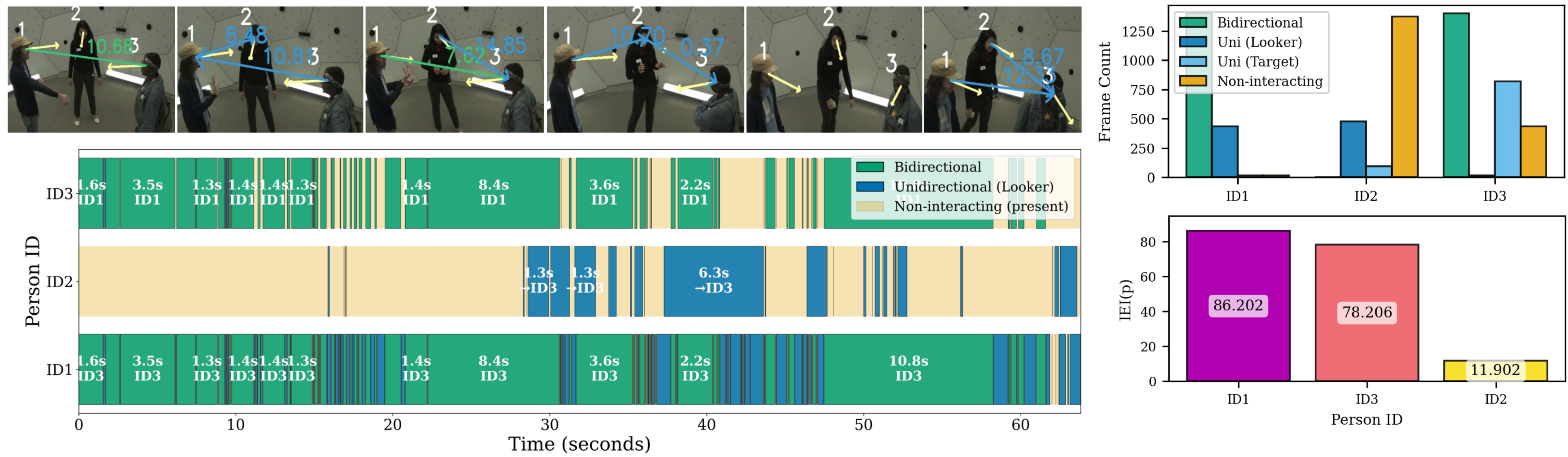}
         \caption{}
         \label{fig:sub2}
     \end{subfigure}
    \caption{ 
    Visualization tool on a sequence from Pexels (Fig. \ref{fig:sub1}) and CMU (Fig. \ref{fig:sub2}). In each of them, on the left we report the sequential frames with estimated head directions and interaction states (top), and the interaction states over time for each person with partner IDs (bottom). On the right, we report the corresponding individual interaction analysis (top: frame level interaction counts; bottom: IEI($p$) for each person).
    }
    \label{fig:main}
\end{figure*}

\begin{figure}[ht]
\centering
\includegraphics[height=2cm]{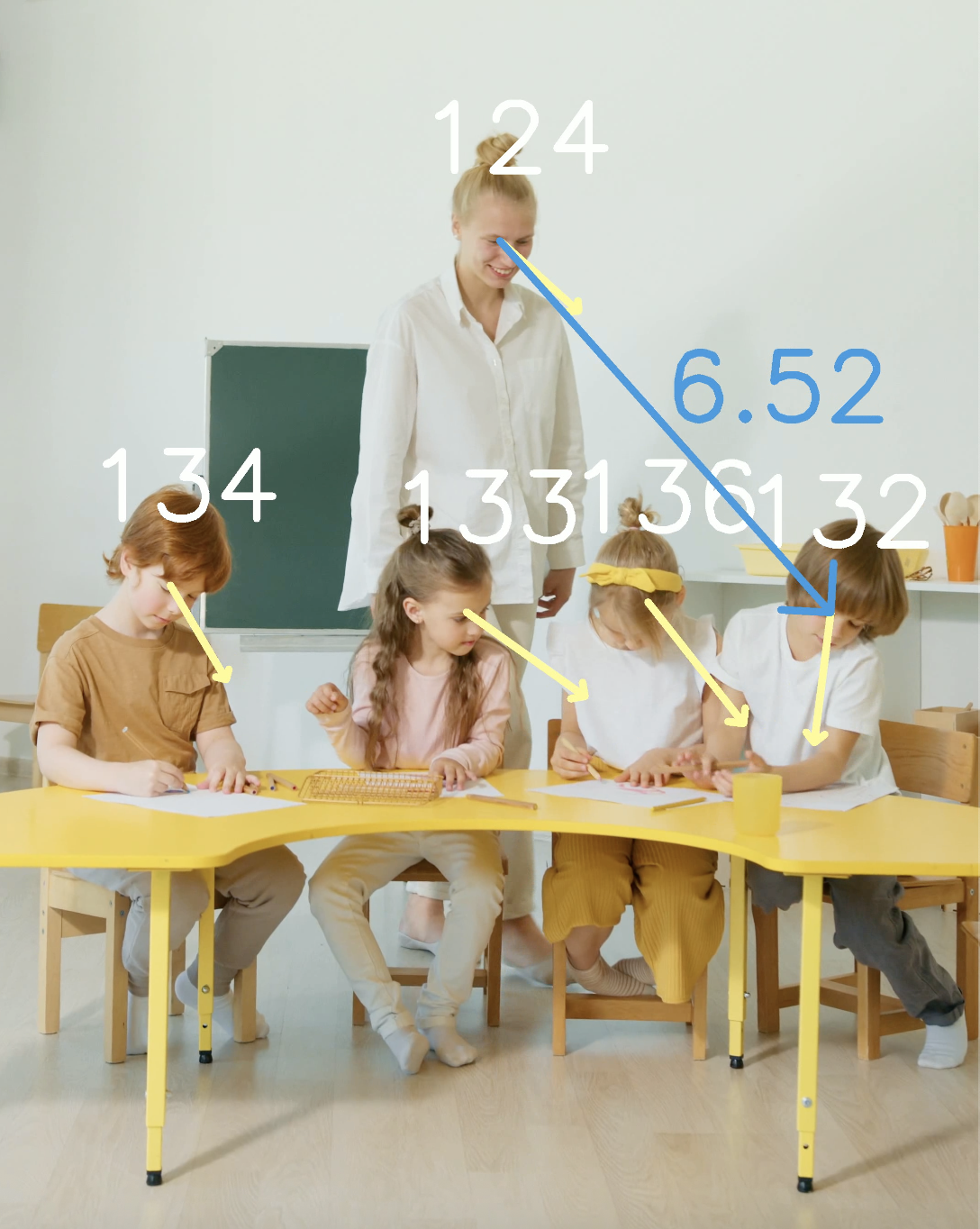}%
\quad
\includegraphics[height=2cm]{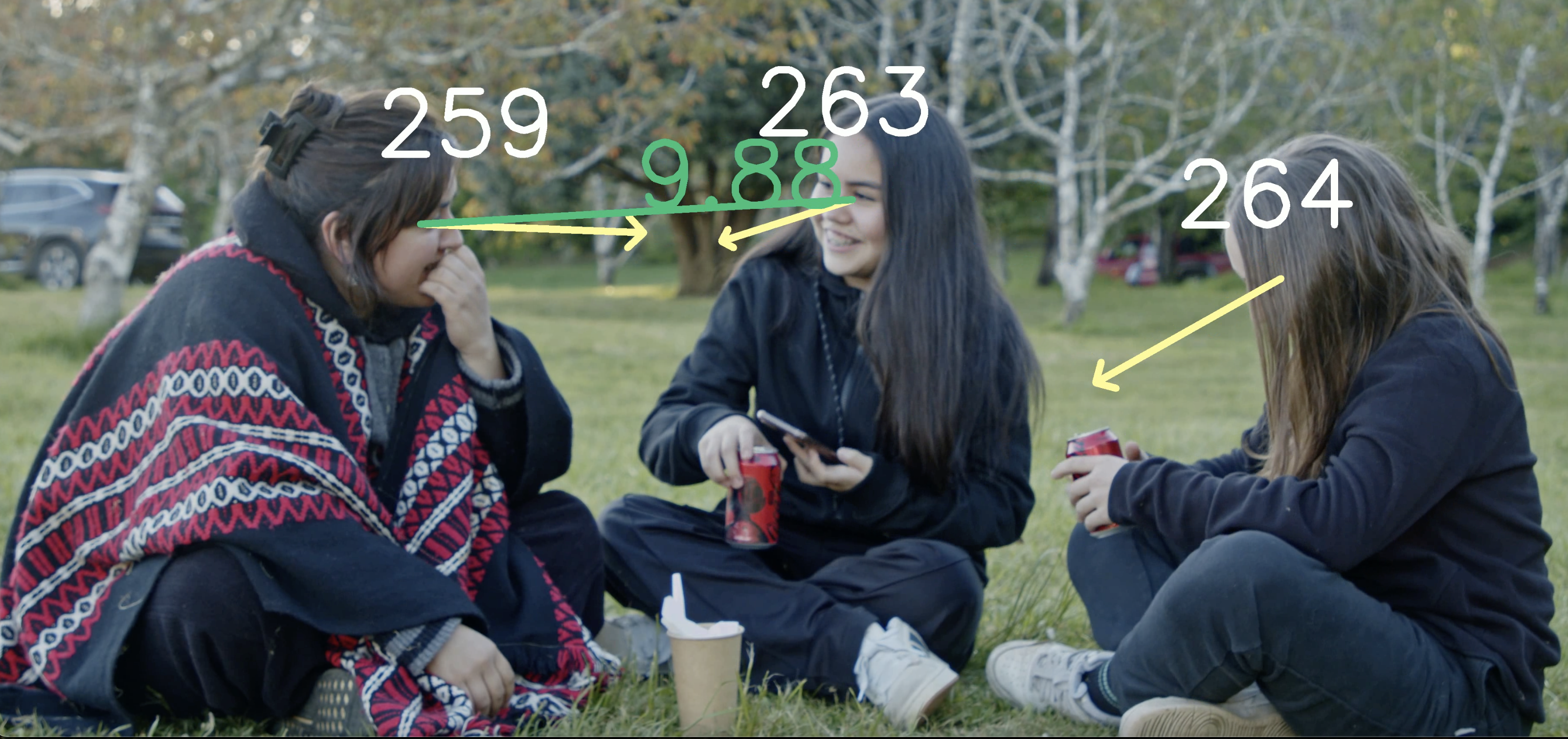}%
\quad
\includegraphics[height=2cm]{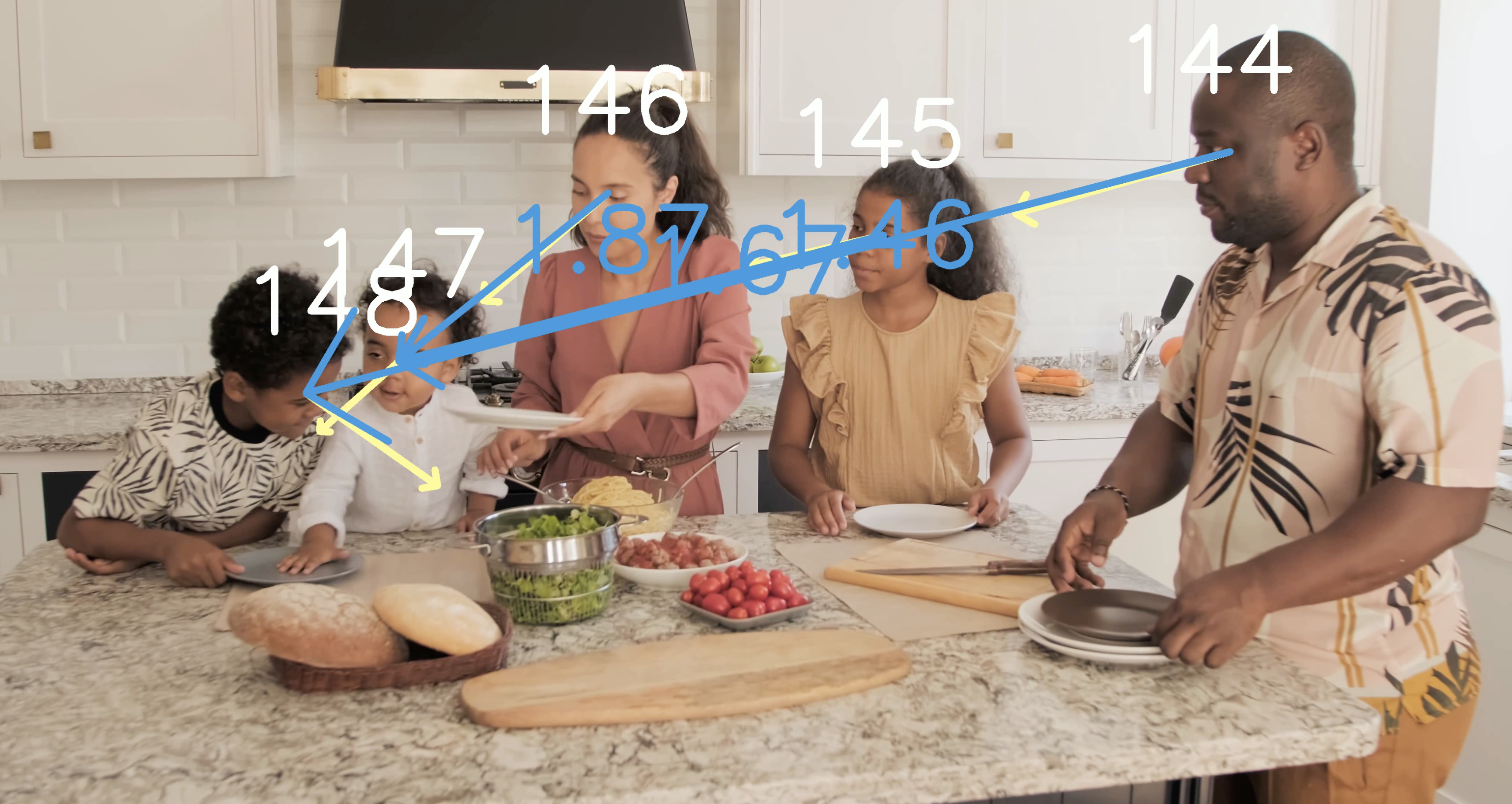}%
\caption{Examples from the Pexels sequences with different interaction scenarios. From left, GES is 31.8, 68.7, and 80.6.}
\label{fig:scenesPexels}
\end{figure}

\noindent {\bf Qualitative results (Fig. ~\ref{fig:main}). }
Each person is assigned a unique ID and their head direction is indicated by a yellow arrow. Edges represent interactions, with green and blue denoting bidirectional and unidirectional cases. Edge values indicate interaction strength.
The horizontal bar plot shows interaction events over time (left in the figure). Each bar represents a person, with color-coded segments indicating interaction states. Partner IDs and interaction durations (in seconds) are shown above each bar, providing an overview of interaction continuity, frequency, and partner switching.
For a compact summary, we also provide histogram-based visualisations (right in the figure).\\
We report visualization samples from  Pexels (top) and CMU (bottom) datasets.
The Pexel sequence depicts a three-person interaction, two children (ID2 and ID3) sit at a table while a woman (ID1) approaches to clean the hands of a child (ID3). Temporal bars indicate that the children interact only during the first 1.3 seconds, while the woman observes them. Afterward, all individuals remain largely isolated. The resulting GES is 40.9. In Fig.~\ref{fig:sub1}, right, we observe a coherent situation. Frame-level interaction counts confirm that non-interaction dominates for the children, while the woman mainly observes the child with ID3.
The first shows the number of frames per interaction state for each person (top), while the second reports IEI($p$) (bottom), indicating individual engagement. Together, they provide an individual and group interaction dynamics.\\
The CMU example shows a conversation between three peers who are taking turns in the discussion, and for this reason we notice a more fragmented interaction type (Fig. \ref{fig:sub2}). In this second example, the GES amounts to 68.9, showing a higher interaction activities with respect to the previous situation.
More in general, we empirically observed that the GES index nicely reflects the interaction scenarios. In Fig. \ref{fig:scenesPexels} we reports examples from the Pexels sequences. On the left, a teacher is instructing the students, each one engaged in their activity. In this case the GES is 31.8, since only the teacher appears to be interacting (i.e. as looker in unidirectional events) with others. The sequence in the middle is very similar to the one reported in Fig. \ref{fig:sub2} and the GES values are indeed very close. Finally, on the right  a family is engaged in a cooperative activity: their GES is 80.6, showing the higher level of interaction in the scene.   

\section{Conclusion}

%

We presented an interpretable and modular framework for video-based social interaction analysis that follows a human-inspired hierarchical process to infer group engagement from dyadic visual attention cues. Rather than replacing human reasoning with a black-box predictor, the framework provides explicit and interpretable intermediate representations that computationally support it.
It combines state-of-the-art pose and head direction estimation with explicit dyadic interaction modelling to derive interpretable measures of engagement at both the individual and group levels. Experiments demonstrate its ability to detect and classify interaction events, while the resulting engagement indices and visualizations make the decision process transparent, enabling teachers, caregivers, and social workers to understand interaction dynamics instead of receiving only engagement scores.\
The main limitation of the current approach lies in its reliance on 2D pose estimation, which may introduce ambiguities due to the projection of three-dimensional scenes onto the image plane. Future work will extend the framework by incorporating 3D human representations, multimodal cues, speaker attribution, and interaction patterns beyond dyadic relationships. A quantitative evaluation of the engagement indexes is also a future objective.

\section*{Acknowledgements}
This work was supported by the European Union — NextGenerationEU and by the Ministry of University and Research (MUR), National Recovery and Resilience Plan (NRRP), Mission 4, Component 2, Investment 1.5, project “RAISE — Robotics and AI for Socio-economic Empowerment” (ECS00000035) We acknowledge the financial support from PNRR MUR Project PE0000013 "Future Artificial Intelligence Research (FAIR)", funded by the European Union – NextGenerationEU, CUP J33C24000430007. This research was partially funded by the European Union - Horizon Europe project “AIRCARE - AI-augmented Robotics for CAncer point of caRE” (101137426), CUP J53C24001360006. 

%
%
\bibliographystyle{splncs04}
\bibliography{main}
\end{document}